# Robust X-ray Sparse-view Phase Tomography via Hierarchical Synthesis Convolutional Neural Networks

Ziling Wu, Abdulaziz Alorf, Ting Yang, Ling Li, Yunhui Zhu

*Abstract*— Convolutional Neural Networks (CNN) based image reconstruction methods have been intensely used for X-ray computed tomography (CT) reconstruction applications. Despite great success, good performance of this data-based approach critically relies on a representative big training data set and a dense convoluted deep network. The indiscriminating convolution connections over all dense layers could be prone to over-fitting, where sampling biases are wrongly integrated as features for the reconstruction. In this paper, we report a robust hierarchical synthesis reconstruction approach, where training data is pre-processed to separate the information on the domains where sampling biases are suspected. These split bands are then trained separately and combined successively through a hierarchical synthesis network. We apply the hierarchical synthesis reconstruction for two important and classical tomography reconstruction scenarios: the spares-view reconstruction and the phase reconstruction. Our simulated and experimental results show that comparable or improved performances are achieved with a dramatic reduction of network complexity and computational cost. This method can be generalized to a wide range of applications including material characterization, in-vivo monitoring and dynamic 4D imaging.

## I. INTRODUCTION

X-ray computed tomography (CT) has been widely used for non-destructive 3D inspection of the internal structures of materials and bio-samples. A 3-D object can be reconstructed from a series of projection images at different viewing angles (also known as the sinogram) using the inverse Radon transform. Unfortunately, the inverse problem of Radon transform is ill-posed, and introduce various reconstruction artifacts including the ring effects, beam hardening, diffraction blurring, and photon noise. These artifacts and aberration effects are usually highly nonuniform. For example, the spatial spectral response of Radon transform severely decays at the high end, which makes the inversion prone to noise amplification at the high frequency region[1],[2]. The high spatial frequency insensitivity of CT images deteriorates with reduced projection angles, and is considered as a major obstacle to scanning speed[3]. Spectrum aberration is also found in phase tomography, where phase induced diffraction emphasizes the high-spatial frequency components and ignores the low-frequency information[4]. On the other hand, Poisson's noise is highly sensitive to the intensity of photon flux. The signal-to-noise ratio is dramatically reduced in dimmer places, consequently[5].

Efforts have been devoted to improving the quality of tomographic image reconstruction since its invention in the 1970s. Before the age of data science, designed spectral filters were used in the filtered back-projection (FBP) method to balance the response curve across high and low spectral regions, and reduce the overall noise. Typical methods include structural adaptive filtering[6], penalized weighted least-square[7], and bilateral filtering[8]. However, filtering generally introduces high spatial frequency information loss. Later, iterative optimization methods were developed to enable reconstruction in the preferred sparse representation of the object, resulting in improved reconstruction with insufficient acquisition. However, these image priors are not always available or easily represented in the analytical form of a sparsity constraint. Besides, such method remains computationally expensive because of the nature of iterative optimization, which limits its widespread applications.

Recent developments in machine learning approaches of convolutional neural networks(CNN) and deep learning techniques[9] have enabled a data-based approach, where the sparsity basis, or more general prior feature information, is learned from sample data[10]. Given a set of corrupted tomography reconstructions and the corresponding ground truth, usually obtained from idealized acquisition conditions, the technique can provide an end-to-end solution that converts future corrupted reconstruction to an improved estimate. A series of rapid developments in machine learning methods over the past few years have come up with various neural network structures for tomographic image reconstruction. In particular, deep learning networks with a large number of layers and connections across different spatial resolutions provide a better recognition of an object's sparsity features on different scales. Successful examples includes residual learning networks using U-Net[11] and mixed-scale dense CNN approach[12].

The dense and convoluted connections in the deep learning architecture generates a large number of mediate images and

---

This work was supported in part by the NSF funding CMMI-1825646.

Ziling Wu, Abdulaziz Alorf, and Yunhui Zhu are with the Bradley Department of Electrical & Computer Engineering, Virginia Tech Blacksburg, VA 24060, USA(e-mail zilingwu@vt.edu).

Ling Li and Ting Yang are with the Department of Mechanical Engineering, Virginia Tech Blacksburg, VA 24060, USA(e-mail lingli@vt.edu).

millions of trainable parameters. While providing impressive results for reconstruction image enhancement, the performance of the CNN-based reconstruction methods has suffered from two problems: the sensitivity to the bias in training data and the lengthy training time. It has been recently confirmed that the imaging enhancement performance is negatively influenced by the biases in the training data, which are not systematically learned and identified by the CNN[13]. For example, the lack of representation of high frequency components caused by deficient sampling, which can be resultant from the pinkish spectrum of the sparse-view CT reconstruction, can be wrongly interpreted as a feature of the object. Such overfitting errors will limit the scope of applicable data, reducing the method's usefulness in practical applications. In addition, deep networks involving dense multi-scale layers are computationally expensive to train. A practical imaging reconstruction method needs to be able to treat systematic bias for a more universally adaptive solution with less computational time.

In this paper, we propose a novel CNN-based imaging reconstruction method that introduces a split-and-combine learning to address for the possible sampling biases. The training samples are pre-processed to split the information into different bands on domains that are prone to processing biases, and then combined in a hierarchical synthesis network. In our application to tomographic reconstruction, the splitting pre-processing is implemented in both the spatial spectral and the intensity domains. We train these split bands in a hierarchical synthesis network, which ensures that different spatial and intensity components are rebalanced correctly in the final reconstruction. Our main contributions are as follows:

- A novel strategy to split the training data on domains of potential biases. This separation pre-processing allows for a guided learning, where physical insights of reconstruction aberration and noise can be integrated into the learning process to avoid overfitting.

- A hierarchical synthesis network that is more adaptive to data pool with multiple sampling biases. The synthesis stages of the network enable the rebalance of the data against different sampling biases one by one. As a result, the learning scheme is more robust against sampling bias and aberrations introduced in the forward modeling.

- An efficient network configuration that works with multi-band information without introducing dense inter-band connections. Much less computational efforts are required compared to dense connected DNN approaches.

- Application to both sparse-view X-ray tomography and phase tomography. The proposed method proves to correct for the spectral and intensity biases in both scenarios.

- Successive experimental sparse-view X-ray phase tomography reconstruction. We have demonstrated high-fidelity X-ray phase tomography reconstruction with a very sparse number of projection angles (75, as compared to the Nyquist requirement of 2800). Comparing to other state-of-the-art methods including framing U-Net[14] and image-to-image translation via conditional generative adversarial network (CAN)[15], we achieve better performance with a significantly reduced the computational time.

## II. BACKGROUND KNOWLEDGE AND RELATED WORKS

This section provides basic knowledge related to X-ray tomographic reconstruction and physical principles related to our applications: the spares-view tomography reconstruction and the phase tomography reconstruction.

### A. Computed Tomography

X-ray Computed tomography has been widely used for non-destructive 3D view of the internal structures for the past decades. **Fig. 1(a)** illustrates the schematic diagrams of X-ray CT with parallel X-ray beam. The forward model of CT is mathematically formulated as the Radon transform $R$, where projection measurement $g$ is obtained by the integral along each projection line $l$ of the object function $f$

$$g(l, \theta) = R(l, \theta)[f(\boldsymbol{r})] \equiv \int_l f(\boldsymbol{r}) d\boldsymbol{r}, \quad (1)$$

where $\theta$ is the projection angle. For attenuation-based CT applications, $f(\boldsymbol{r})$ is the X-ray linear attenuation coefficients at different voxels of the object. A CT reconstruction problem is formulated as retrieving the unknown function $f$ based on the observed sinogram $g(l, \theta)$.

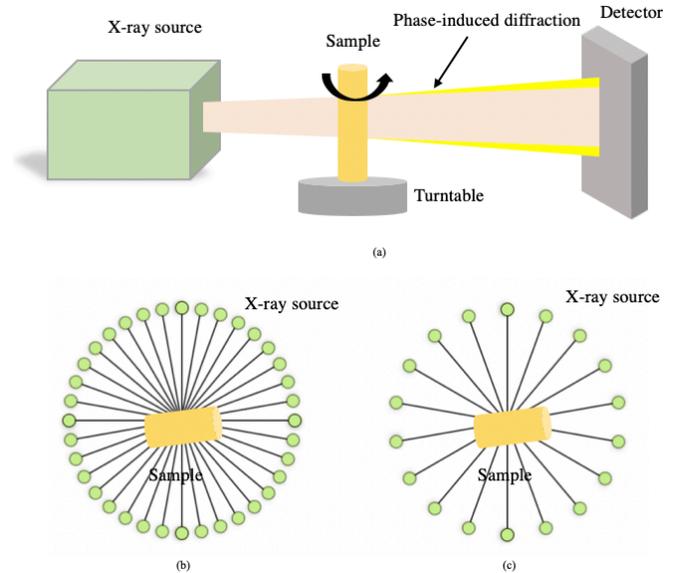

Fig. 1. Schematic illustration of (a) X-ray CT configuration, sideview (b) full-view CT, and (c) sparse-view CT, top view.

### B. Under-sampled CT reconstruction

Driven by the demand to reduce the X-ray radiation dose and scanning time in medical and industrial applications, people have been exploring CT reconstruction based on insufficient acquisition, which are prone to artifacts and noise. For example, low-dose CT, which is implemented with a decreased X-ray illumination, or equivalently, a reduced exposure time [16][17], is prone to Poisson's noise as result of insufficient photon counts. On the other hand, spare-view CT is obtained with a reduced number of projection images, as illustrated in **Fig. 1(c)**. This acquisition does not satisfy the Nyquist sampling rule[18]. According to Fourier slice theorem, the insufficient angle sampling introduces under-sampling in the outer rim of the Fourier spectrum, causing streaking artifacts[19].

Many studies[10], [20]–[26]) have proposed improving under-sampled CT reconstruction by compressive sensing, i.e., including a constraint function $\varphi$ that carries the prior information of the object. The improved reconstruction $\hat{f}$ is obtained via the joint optimization

$$\hat{f} = \text{argmin}\{\|Rf - g\|^2 + \alpha\varphi(f)\} \qquad (2)$$

where $\alpha$ is the regularization parameter controlling the weight between the fidelity term and the constraint regularizer. Various prior regularizers have been applied to CT reconstruction of different objects of interests, including total variation (TV) and its variants([20]–[23]), non-local means[24], dictionary learning[25], low-rank and its variants[26][10], and etc., but they generally suffer from long computational time on iterative optimization[27], [28]. There is also limitation regarding the inefficiency to represent global features and constructing the regularizer for arbitrary objects [29].

Recently, several CNN architectures obtained impressive results for under-sampled CT reconstructions. In particular, multi-scale deep learning methods provide a best performance by capturing features across various spatial scales. To name a few, Han et al[10] proposed a U-Net[30] structured architecture with residual learning to remove the artifacts in sparse-angle reconstruction CT image. Lee et al.[31] proposed CNN to interpolate missing data of sinogram for sparse-view CT by combining with residual learning for better convergence and patch-wisely training. Isola et al.[15] proposed image-to-image translation using a conditional generative adversarial network (CAN) that learns the mapping between an input/corrupted image and an output/clean image using a training set of aligned image pairs. Kang et al.[32] proposed a combination of different frequency components through directional wavelets in the deep convolutional neural network for low-dose CT reconstruction. Zhang et al.[33] combined the DenseNet with deconvolution for sparse-view CT reconstruction. Pelt et al.[12] proposed a mixed-scale dense CNN for image segmentation which applies dilated convolutions to capture features at different image scales and densely connected all feature maps with each other.

Although the previous deep learning studies have achieved high quality CT reconstructions, most of them failed to interpret the success with respect to inverse problems, thus cannot be adapted to different biases introduced from the forward modeling. In particular, the existing U-Net architecture does not satisfy the framing condition for non-local basis imposed by deep convolutional framelets, which often result in the emphasis of the low frequency component of the signal (blurring artifacts). Our method, on the other hand, is capable of dealing with biases in multiple domains in an explicit way, therefore can be adaptive to various acquisition conditions in a robust manner.

## C. X-ray phase imaging and phase tomography

While X-ray imaging and CT have been predominately based on the absorption of X-ray, it is recently found that phase signals, especially from soft materials such as biological tissues, provide an enhanced contrast. This is because the phase delay of the X-rays introduced by the soft materials could be three order of magnitude larger than its attenuation [34][35]. In the past several years, several quantitative X-ray phase imaging techniques have been developed, including propagation-based [36]-[38], analyzer-based[39], grating-based[40]-[42] approaches, as well as far-field ptychography[15].

In particular, transport of intensity based phase imaging[34], [35], [45] has gained popularity due to its simplicity and low requirement on source coherence. Based on Fresnel diffraction, an object with a phase profile of $\phi(x, y)$, will introduce a measurable transport of intensity $I$ as light wave propagates along the $z$ axis, expressed by

$$\left.\frac{\partial I(x, y; z)}{\partial z}\right|_{z=0} = -\frac{1}{k}\nabla_\perp \cdot (I(x, y; z = 0)\nabla_\perp\phi(x, y)). \qquad (3)$$

Here $k = 2\pi/\lambda$ is the wavenumber, $(x, y)$ is the position vector in the transverse plane perpendicular to the optical axis $z$, and $\nabla_\perp = \partial/\partial x + \partial/\partial y$ is the gradient operator in the transverse plane. Equation (3) can be simplified for a single-material object, in which the duality ratio $\gamma$ of the refractive index and attenuation coefficient is constant throughout[10], [45]. The phase profile $\phi(x, y)$ is then retrieved from a single shot measurement of $I(x, y; z)$ by

$$\phi = \exp\left(-2\gamma\right)F^{-1}\left(\frac{1}{\pi\gamma z\lambda k^2}F[I(z)]\right), \qquad (4)$$

where $F$ and $F^{-1}$ denote as Fourier and inverse Fourier transform operators, respectively.

Recent development of X-ray phase tomography is based on sinogram of retrieved projection phase profiles[38]. In additional to the common CT artifacts, phase tomography also suffers from diffraction induced low-frequency noise amplification. Similarly, iterative regularization optimization [46], [47], in particular total variation-based compressive priors[47], have been applied for phase tomography enhancement. A few machine learning approaches without explicit prior information has been introduced for quantitative phase retrieval and phase tomography. Nguyen et al.[48] applied a single-stage DNN to optical tomography for reconstructing 3D phase distribution. Goy et al.[49] recently developed high-resolution limited angle phase tomography reconstruction with deep neural networks to establish the end-to-end approach with synthetic training data to emulate X-rays phase tomography. To our best knowledge, there is no CNN based method developed for X-ray phase tomography with real X-ray data yet. Moreover, none of the previous approaches addresses the specific image corruption and sampling biases induced by phase imaging and tomography.

## D. The splitting-and-combination approach

It is a common strategy to split the information in different domains to address for potential bias in imaging [50]-[54] and computer vision communities[55]. Many works have been inspired by the HiLo microscopy[50], which improves optical imaging resolution by splitting spatial spectral information into low and high frequency bands[51]. Similar HiLo splitting has been implemented as pre-modulation in several machine learning approaches, including DualCNN[52], learning to synthesize network[53] and multi-resolution phase recovery[54]. However, these methods have been predominately applied in the spatial spectral or resolution domain. Here, we propose a hierarchical synthesis network architecture to combine information split in multiple domains, and thereby extend the scope of application. For the specific application of sparse-view phase tomographic reconstruction, we split based on the observed biases in two of the spatial spectral and intensity domains.

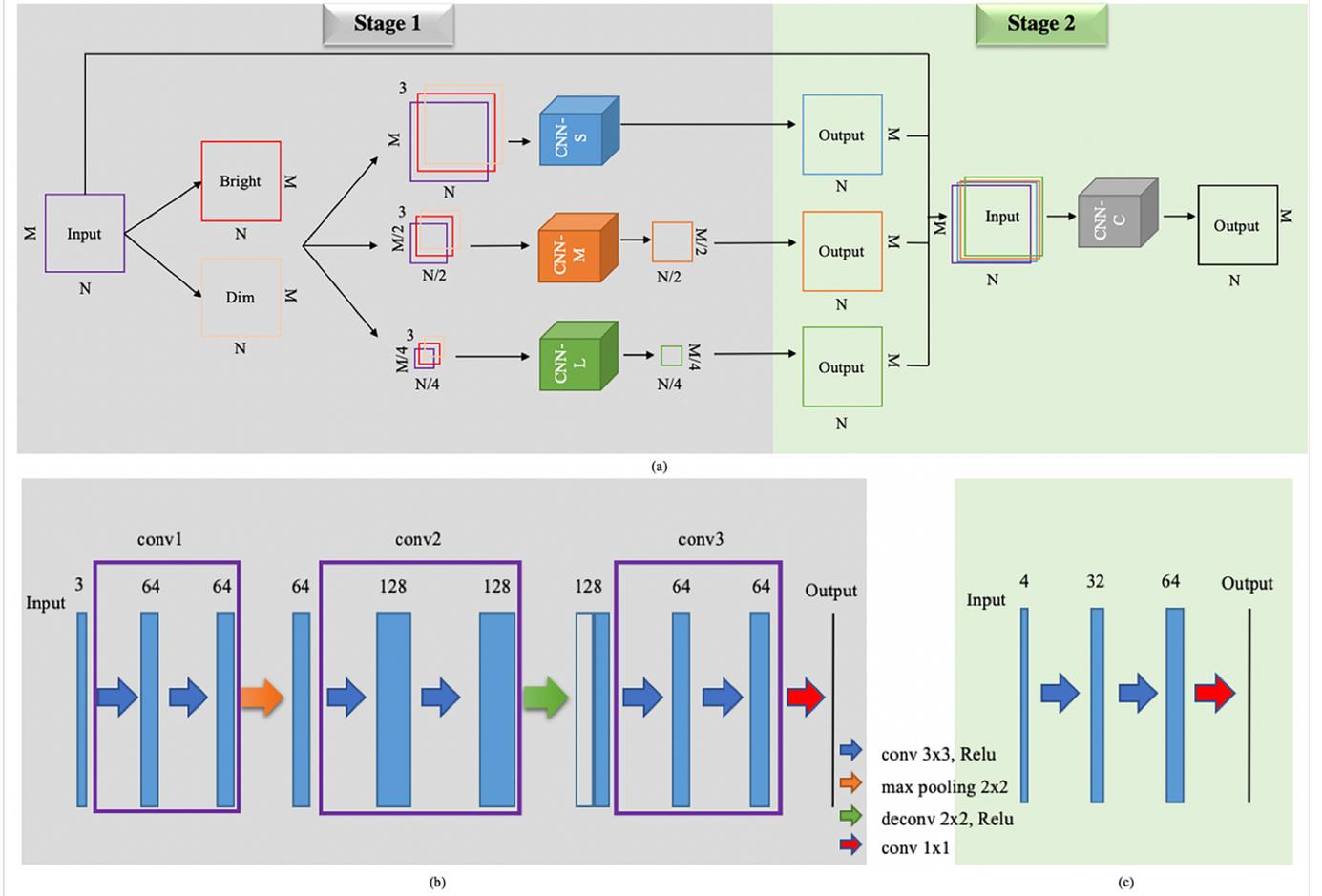

Fig. 2. (a) Proposed overall network architecture of Hierarchical synthesis of CNNs(HSCNN). Detailed network construction for feature fusion are shown in (b) for the first stage and in and (c) for the second stage.

## III. PROPOSED HIERARCHICAL SYNTHESIS CNN.

### A. Hierarchical synthesis of CNNs (HSCNN)

The proposed method involves a hierarchical split-and-combine process for multi-domain bias correction. We choose to separate the training data in both the spatial spectral domain and the intensity domain, where non-uniformity in the forward transfer functions are observed in tomographic reconstructions. As shown in **Fig. 2(a)**, the corrupted tomographic reconstruction images are first split by intensity. The bright band and the dim band are obtained by filtering pixels above and below the average intensity, respectively. These two intensity variations, together with the original inputs, are then successively split again into three spatial channels via low-pass filters on different resolution scales. As shown in the **Fig. 2(a)**, a total of 9 copies of the training data are generated with different spectral and intensity distributions. Next, these split sample data are fed into a hierarchical synthesis learning network. The first training stage combines the intensity variations to generate an output of feature maps that matches the ground truth expectations on each of the three different resolution scales. These multi-scale feature maps together with original input are then used as the input for a second training stage, which combines the spatial spectral components for the final reconstruction. Each of the separate CNN networks within the same stage can be trained in parallel, allowing for a fast learning of features with different scales and intensities. This hierarchical architecture can be easily extended to include more domain stages and more bands within each stage.

### B. Architecture of CNNs for feature fusion on each stage

The training nodes in stage 1 use an architecture as shown in **Fig. 2(b)**. Three convolutional blocks (denoted as conv1, conv2 and conv3), each consists of two convolution layers and two rectified linear units (ReLU)[56] layers. The number of feature maps is set to 64. A max pooling layer is placed between conv1 and conv2 for downsampling. The size of kernel is defined to $2 \times 2$ and the stride is set to 2 in the pooling layer. A deconvolutional layer is placed between conv2 and conv3 for upsampling. The size of kernel is defined to 2×2 and the stride is set to 2 in the deconvolutional layer. At the end, a 1x1 convolutional layer followed with softmax and pixel classification layers are added to predict the categorical label for each image pixel.

The second training stage uses an architecture shown in **Fig. 2(c)**. Three scale maps and the original input image are first concatenated into a 4-channel image. Then three convolutional blocks are used to fuse the feature maps and predict the enhancement. For the former two convolutional blocks, we set the number of feature maps as 32 and 64, respectively, each convolutional block contains one convolutional layer and one ReLU layer. We use one convolutional kernel to convert the features to the final result.

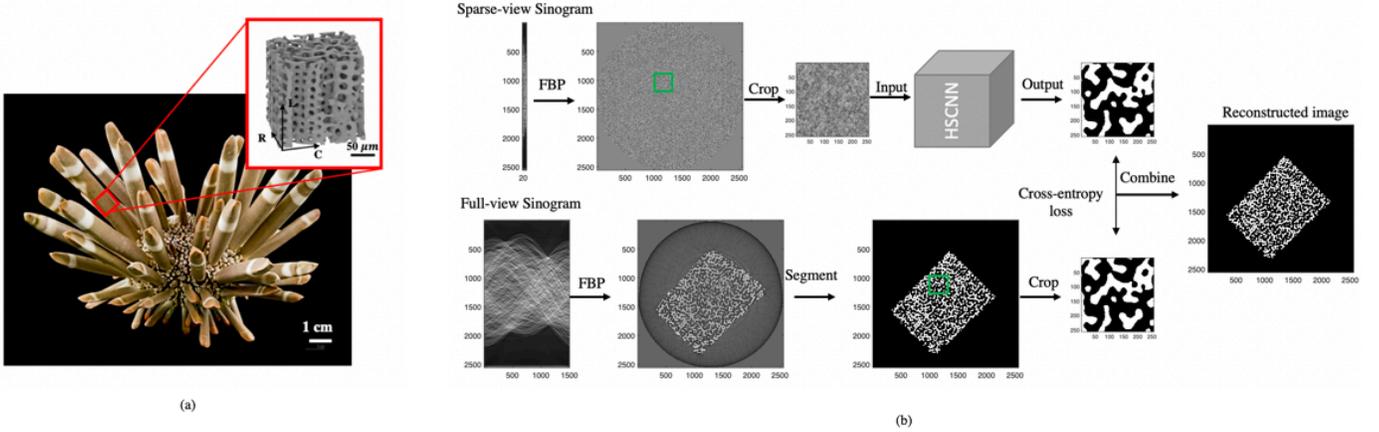

Fig. 3. (a) Photography of the sea urchin spine. The insert shows a zoomed in 3-D cellular structure. (b) Processing flowchart.

## IV. SIMULATION AND EXPERIMENTAL IMPLEMENTATION

We validate our proposed method on a sparse-view phase tomography for a biological sample in both simulation and experiment. The implementation flowchart is shown in **Fig. 3.**

### A. Experiment setup

We chose to use the biological sample of sea urchin spine [57][58] as our imaging object, which consists of bio-ceramic cellular networks, as shown in **Fig. 3(a).** The complex 3D microscopic structure with a single-material composition made this material system an excellent model for duality phase tomography demonstration. The samples were imaged with the synchrotron-based X- ray μ-CT system, at the beamline 2-BM at the Advanced Photon Source of Argonne National Laboratory. The tomography measurement setup was conducted with a monochromatic beam (27.4 KeV). The propagation distance from the sample to the detector plane was 60 mm, allowing for the recording of the phase induced diffraction signals. Each projection image contains 2560x2560 pixels with an isotropic pixel size of 0.65 μm.

A full tomography sinogram $g$ consisted of 1500 projection images is acquired over a 180-degree sample rotation with the rotation speed of 0.5 degree/second. Two sparse-view tomography scans $g_1$ and $g_2$ are implemented with 75 and 50 projections images, evenly distributed over the 180 degree at the speed of 10 degree/second.

### B. Data acquisition

#### 1) Ground truth $\hat{s}$

The ground truth is achieved by phase tomography reconstruction from the full scan sinogram $g$. Quantitative phase retrieval is first obtained via inverse duality TIE in Eq. (4) for each projection image, which is followed by a tomography reconstruction using a Fourier grid reconstruction algorithm implemented in the open source software *Tomopy*[59]. After that, intensity filtering is implemented based on maximum of likelihood estimation[60] to yield a refined results shown in **Fig. 3(b)**. A total volume of 2560×2560×2160 voxels are reconstructed, from which a small fraction is randomly cropped and used for training and testing.

#### 2) Input $s$

Several sets of corrupted tomography reconstructions are generated as the input training data, including two different simulated reconstructions and two sparse-view experimental reconstructions based on $g_1$ and $g_2$.

A phase diffraction measurement is simulated based on the structure $\hat{s}$. Projection intensity measurement is simulated via Eq. (3), where wavelength $\lambda$ of X-ray is set to 0.045 nm and the distance of propagation is set to 60 mm. The material duality ratio is set as 500 for a typical soft material. We also add a white Gaussian noise to the sinograms with an averaged SNR of 20dB. These sinograms are reconstructed through the FBP algorithm and used as corrupted input $s_p$.

We then simulate the attenuation-based projection measurement for a sparse-view acquisition based on Eq. (1).

We simulate for evenly distributed 50-view sinograms. Reconstructions via the FBP method are used as corrupted inputs $s_s$

Finally, the sparse-view experimental phase tomography sinogram $g_1$ and $g_2$ are reconstructed via the Fourier grid reconstruction algorithm implemented in the open source software *Tomopy* and used as corrupted input $s_1$ and $s_2$.

Each reconstruction $s$ contains a total volume of 2560×2560×2160 voxels.

### C. Network training

We train our network in a patch-by-patch manner. We crop 1394 patches with the size of 256×256 pixels randomly from each input $s$ and corresponding ground truth $\hat{s}$ for training, and another 598 patch pairs for testing. The trained network is then used to enhance the untrained slices by translate each patch in the new slice into the enhanced output. These output patches are stitched accordingly to retrieve the whole image reconstruction result, as shown in **Fig. 3(b).**

During training stage, the sigmoid cross-entropy is used as the loss function to train all neural networks mentioned above, which is denoted as the averaged pixel-wise cross entropy loss between output and the ground truth. If $K$ is the number of classes, $t_{ijk}$ is the indicator that the $(i,j)^{th}$ pixel belongs to the $k^{th}$ class, and $y_{ijk}$ is the output for $(i,j)^{th}$ pixel for class $k$, then the loss function is defined as:

$$J = -\frac{1}{MN}\sum_{k=0}^{K}\sum_{i=1}^{M}\sum_{j=1}^{N} t_{ijk}\log\left(y_{ijk}\right) \quad (5)$$

where we used $K = 2$ for binary correlation. Our method is implemented on MATLAB. The machine used for our experiments is a PC with Intel Core i7-6700K 4.0-GHz CPU, 32-GB RAM, GeForce GTX 960 18GB GPU. All images are stored on SSD, which accelerates reading speed. During each training phase, the Adam optimization method was used to train model with a mini-batch of 32 image patches for each iteration. The learning rate was selected to be $1 \times 10^{-4}$. To avoid over-fitting, L2 regularization term is added with the weight of 0.0005. A typical training time is around 30 minutes.

### D. Image Metrics

For a quantitative assessment, we use two image metrics, specifically the peak signal-to-noise ratio (PSNR) and structure similarity index (SSIM). PSNR is defined in terms of the mean square error (MSE)

$$MSE = \frac{1}{MN}\sum_{i=1}^{M}\sum_{j=1}^{N}[Y_{ij} - X_{ij}] \quad (6)$$

where Y is the target image ($\hat{s}$) and X is the translated sparse-view reconstructed image. PSNR is expressed by

$$PSNR = 10\log_{10}(\frac{MAX_Y^2}{MSE}) \quad (7)$$

where $MAX_Y$ is the maximum value of image Y. The PSNR value approaches infinity as the MSE approaches zero; a higher PSNR value indicates a higher image quality.

The SSIM is a well-known quality metric which measures the similarity between two images[61]. SSIM is considered to be correlated with the quality perception of the human visual system (HVS). Suppose $x$ and $y$ are two image blocks of image

$X$ and $Y$, the SSIM index is defined as:

$$SSIM(x,y) = \frac{(2\mu_X\mu_Y+c_1)(2\sigma_{xy}+c_2)}{(\mu_X^2+\mu_Y^2+c_1)(\sigma_X^2+\sigma_Y^2+c_2)} \quad (8)$$

where $\mu$ and $\sigma$ measure the mean and standard deviation of the two image blocks $x$ and $y$. The small positive constants $c_1$ and $c_2$ are used to avoid a null denominator. The block size is typically 8×8, and the final SSIM value between X and Y is the averaged SSIM of all blocks.

## V. RESULTS AND DISCUSSION

We show the results for the simulation and experimental tomography reconstructions and demonstrate the robustness of the method in this section. We compare our method with several main-stream and up-to-date deep learning method and show that better or comparable results are obtained with dramatically improved speed. Averaged PSNR and SSIM values are presented for comparison.

### A. Simulation: phase tomography reconstruction.

The reconstruction results for the simulated phase tomography are shown in **Fig. 4** for one slice outside the training and testing sampling scope. As demonstrated in this result, splitting-and-combining in either the intensity (column (c)) or spectral domain (column (d)) prove to significantly improve the reconstruction quality, as compared to the FBP reconstruction $s_p$ shown in column (b). However, we can still observe clear hollow centers in column (c), resulting from failure to correctly boost the low-spectral component in the learning. At the same time, residual scattered noises are presented in column (d). It is only through the implementing of the hierarchical synthesis network in both domains that we achieve a best result (column 5), which corrects for both the edge effect residuals and the artifacts associated with the bright spots.

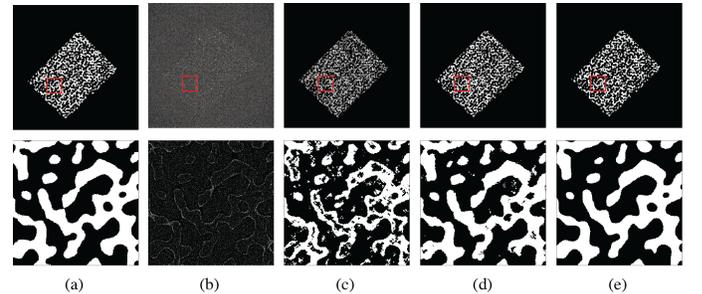

Fig. 4. Reconstruction results of phase tomography simulation: (a) the ground truth (GT) $\hat{s}$ , (b) corrupted reconstruction with FBP $s_p$, HS-CNN learning results for splitting (c) in the intensity domain only, (d) in the spectral domain only and (e) in both domains hierarchically. The second row shows the zoomed-in images from the red boxes.

### B. Simulation: Sparse-view CT reconstruction

The results for the simulated sparse-view CT with 50 projection angles are shown in **Fig. 5**. Serious streaking artifacts in the poorly reconstructed result $s_s$ are corrected with the HSCNN. Again, splitting in a single domain provides less favorable results. In particular, the split-in-intensity-only reconstruction in column (c) demonstrate fussy edges, which is inherited from the high-spatial frequency loss in the sparse-view CT. The spectral-only reconstruction in column (d) provides a reasonable result as expected, since no intensity-related noise is introduced in this simulation. The best reconstruction is obtained by the HSCNN, which yields a highest SSIM of 0.9242 and PSNR of 17.5324.

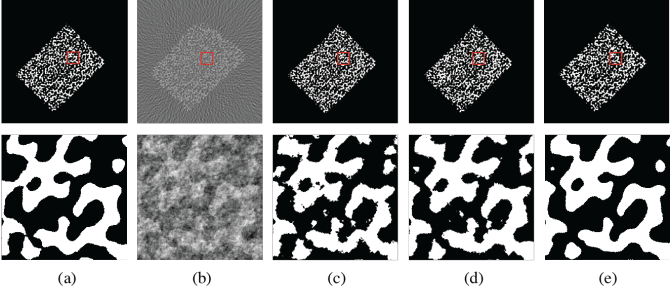

Fig. 5. Reconstruction results of sparse-view tomography simulation with 50 angles. (a) the ground truth $\hat{s}$, (b) FBP results, (c) HSCNN in intensity domain only, (d) HSCNN in spectral domain only and (e) the results of our hierarchical synthesis method. The second row shows the zoomed-in images from the red boxes.

### C. Experiment: Sparse-view phase tomography reconstruction

The sparse-view phase tomography reconstruction results based on the tomographic scan $g_1$ are shown in **Fig. 6**. This real experimental data suffers from both the streaking effect and the diffraction edge effect, as well as Poisson's noise. As a result, the FBP-based reconstruction $s_1$ is prone to noise and the loss of spectral information in both the low and the high ends. It is thus not surprising that structures are barely visible from untrained reconstruction in column (b). HSCNN, nevertheless, provides a high-quality reconstruction in column 5 with a SSIM of 0.9170 and a PSNR of 15.5835.

TABLE I: QUANTITATIVE COMPARISON OF SPLITTING ON DIFFERENT DOMAINS

| Methods\Applications | Phase Tomography | Sparse-view CT | Sparse-view phase Tomography |
|---|---|---|---|
| Intensity-only | 0.8552/13.9304 | 0.7607/8.2948 | 0.8527/13.8403 |
| Spectral-only | 0.9202/17.3740 | 0.8895/13.9781 | 0.8894/14.9243 |
| HSCNN | **0.9402/19.2459** | **0.9242/17.5324** | **0.9170/15.5835** |

Both the simulation and experiment results show that it is desirable to use a hierarchical synthesis learning architecture to include multiple bias correction. Quantitative image metrics shown in Table I echoes with the observation.

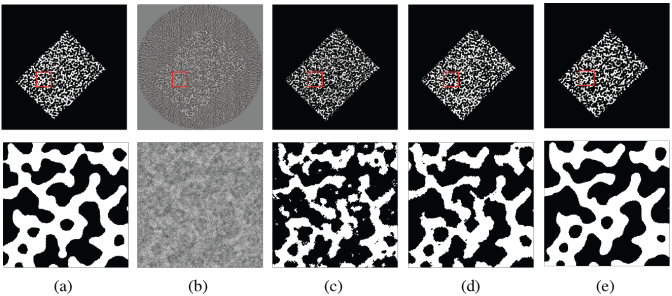

Fig. 6. Reconstruction results of sparse-view phase tomography experiment with 75 angles. (a) the ground truth, (b) FBP results, (c) HSCNN results in intensity domain, (d) HSCNN results in spectral domain, and (e) results in both domains hierarchically. The second row shows the zoomed-in images from the red boxes.

### D. Transfer learning of sparse-view CT reconstruction

We demonstrate the robustness of the training networking in a transfer learning between two different acquisition scenarios (i.e., sparse-view phase CT with different viewing angles). By avoiding overfitting through densely interconnected networks, the trained re-balance synthesis network can be transferred to data with similar spectral aberrations. As shown in **Fig. 7**, the hierarchical synthesis network is first trained by the 75-angle sparse-view CT scan $s_1$, and then directly applied to the 50-angle sparse view scan $s_2$. Similar high-quality reconstructions are achieved with a SSIM of 0.9004 and a PSNR of 14.8891. This is compared with a SSIM of 0.9037 and a PSNR of 14.9893 obtained by reconstructing $s_2$ with a HSCNN network trained by data from $s_2$. **Fig. 7** shows the consistency between transferred and non-transferred results. Both results are also highly correlated to the ground truth.

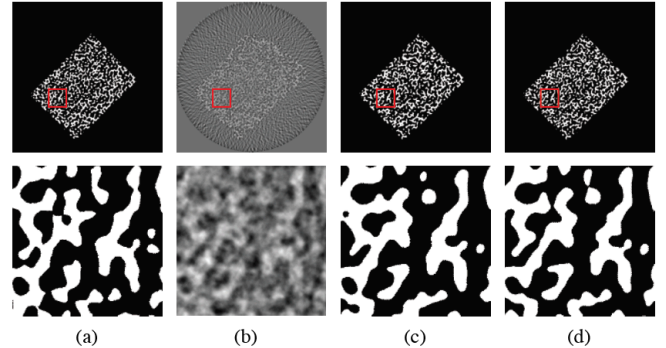

Fig. 7. Transfer learning results. (a) the ground truth, (b) FBP result based on 50-views sinogram, (c) translation of (b) with a HSCNN pre-trained on the 75-view data, (d) translation of (b) with a HSCNN pre-trained on the 50-view data.

### E. Comparison with other methods

We benchmarked our proposed method (HSCNN) against three methods for image quality reconstruction: (1) compressive sensing with TV minimization (2) framing U-Net proposed by Han et al. [14] and (3) conditional adversarial networks(CAN) proposed by Isola et al.[15]

Total variation minimization method has been a widely used compressive sensing imaging reconstruction method. The reconstruction is achieved by iterative optimization of $f$ from the measurement $g_1$ based on Eq. (3). It was implemented on a single slice of $s_1$ for demonstration.

The framing U-Net learns through the mathematical model of deep convolutional framelets. A high spectral emphasis is imposed through framing which improves the high frequency recovery[11]. We use it as an advanced U-Net approach that addresses the spectral bias in CT reconstruction.

The conditional adversarial network has emerged as a general-purpose image-to-image translation solution, which produces output images from conditioned input images by learning the structural loss function through the training pairs.

To enable a fair comparison, the framing U-Net and CAN are trained with the same dataset of $s_1$, both using 1394 corrupted images with the size of 256x256 pixels. The training is implemented on a PC with Python 3.5, Ubuntu Linux 17.10, 128 GB RAM, Intel Core i9 4.6GHz CPU, and a Nvidia TITAN Xp GPU with 12 GB Memory. The training time was about 1 day and 8 hours, respectively.

TABLE II: QUANTITATIVE COMPARISON WITH OTHER METHODS

| Metric\Method | TV | Framing U-net | Image-to-image translation | HSCNN |
|---|---|---|---|---|
| PSNR(dB) | 4.5873 | 3.4158 | **10.8214** | 10.5476 |
| SSIM | 0.0376 | 0.2807 | 0.5978 | **0.6793** |

As shown in Table II and **Fig. 8**, best results are generated by our method (HSCNN), while CAN method achieves visionally similar performance. The framing U-Net result is significantly flawed, while the TV minimization fails to yield a readable figure. Given the simplicity of our proposed method, and the easiness/speed to train and test, our proposed method outperforms all other methods.

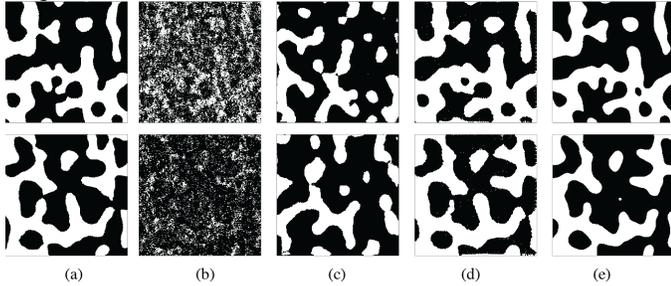

(a)    (b)    (c)    (d)    (e)

Fig. 8. Comparison with other methods. (a) the ground truth are compared with the reconstruction results from (b) TV minimization, (c) Framing U-net, (d) CAN and (e) our method of HSCNN. Two zoomed-in samples are shown for each method.

## VI. CONCLUSION

We have presented a hierarchical synthesis CNN network architecture for sparse-view and phase X-ray tomography image reconstruction. By pre-separating features in the domain with potential biases, a split-and-combine strategy is implemented to correct for the nonuniformity in the forward model. In addition, by using a hierarchical synthesis structure, we are able to fuse multi-band information without introducing dense connections across different bands. Accurate reconstructions are obtained for simulated and experimental sparse-view phase tomography, which outperforms popular alternative approaches including compressive sensing via TV minimization, frame U-net, CAN in term of image quality and computational speed.

Generalizations are possible with biases in more than two domains, and more bands within each domain. The future work will be extending the current framework to include 3D spatial correlation and use it for dynamic 4D imaging.